\documentclass{article} 
\usepackage{iclr2023_conference,times}

\usepackage{amsmath,amsfonts,bm}

\def\eqref#1{equation~\ref{#1}}

\def\1{\bm{1}}

\DeclareMathAlphabet{\mathsfit}{\encodingdefault}{\sfdefault}{m}{sl}
\SetMathAlphabet{\mathsfit}{bold}{\encodingdefault}{\sfdefault}{bx}{n}

\usepackage{hyperref}
\usepackage{url}
\usepackage[pdftex]{graphicx} 

\usepackage{booktabs}   
\usepackage{subcaption} 
\usepackage{array}   
\newcolumntype{L}{>{$}l<{$}} 
\newcolumntype{T}{>{\centering\arraybackslash}m{3cm}}
\title{ML-driven Hardware Cost Model for MLIR}         

\author{Dibyendu Das and Sandya Mannarswamy \\
 Intel \\
 Bangalore, India \\                    
\texttt{\{dibyendu.das,sandya.mannarswamy\}@intel.com} \\         
}

\iclrfinalcopy 
\begin{document}
\maketitle
\begin{abstract}
Compiler optimizations are diverse and have different payoffs for different types of code generated. During early optimization passes, compilers must make predictions for machine-dependent characteristics such as execution unit utilization, number of register spills, latency, throughput etc. to generate better code. Often a hand-written static/analytical hardware cost model is built into the compiler. For example, in LLVM, TTI is used extensively as a surrogate for actual performance. However, the need for more sophisticated and varied predictions has become more pronounced with the development of deep learning compilers which need to optimize dataflow graphs. Such compilers usually employ a much higher level MLIR form as an IR representation before lowering to traditional LLVM-IR. A static/analytical cost model in such a scenario is cumbersome and error prone as the opcodes represent very high level algebraic/arithmetic operations. Hence, we develop a machine learning-based cost model for high-level MLIR which can predict different target variables of interest such as CPU/GPU/xPU utilization, instructions executed, register usage etc. By considering the incoming MLIR as a text input a la NLP models we can apply well-known techniques from modern NLP research to help predict hardware characteristics more accurately. We expect such precise ML-driven hardware cost models to guide our deep learning compiler in graph level optimizations around operator fusion, local memory allocation, kernel scheduling etc. as well as in many kernel-level optimizations such as loop interchange, LICM and unroll. They can also help dynamic runtimes make decisions on whether to incur the cost of recompilation given changing operator shapes or continue using already compiled code. We report early work-in -progress results of developing such models on high-level MLIR representing dataflow graphs emitted by Pytorch/Tensorflow-like frameworks as well as lower-level dialects like affine. We show that these models can provide reasonably good estimates with low error bounds for various hardware characteristics of interest and can be a go-to mechanism for hardware cost modelling in the future. 
\end{abstract}

\section{Introduction}
Of late, considerable strides have been made in applying deep learning (DL) techniques to software engineering itself, including source code assistance, automatic source code generation and in building software tools ~\cite{Le2020}. The emergence of open-source community software development and large code repositories such as GitHub have accelerated interest in applying DL techniques to programming, compiler optimizations, code generation etc. Neural models have been developed for source code ~\cite{Alon2019} and intermediate code representations ~\cite{VenkataKeerthy2020}. ML models have been used for cost prediction and heuristics selection in compiler optimizations ~\cite{Leather2020}, ~\cite{Troffin2021}, ~\cite{HajAli2020}, \cite{Das2020}, \cite{NIC2022}, \cite{Huang2019}. 

\begin{figure}[ht]
  \begin{center}
  \includegraphics[width=0.9\linewidth]{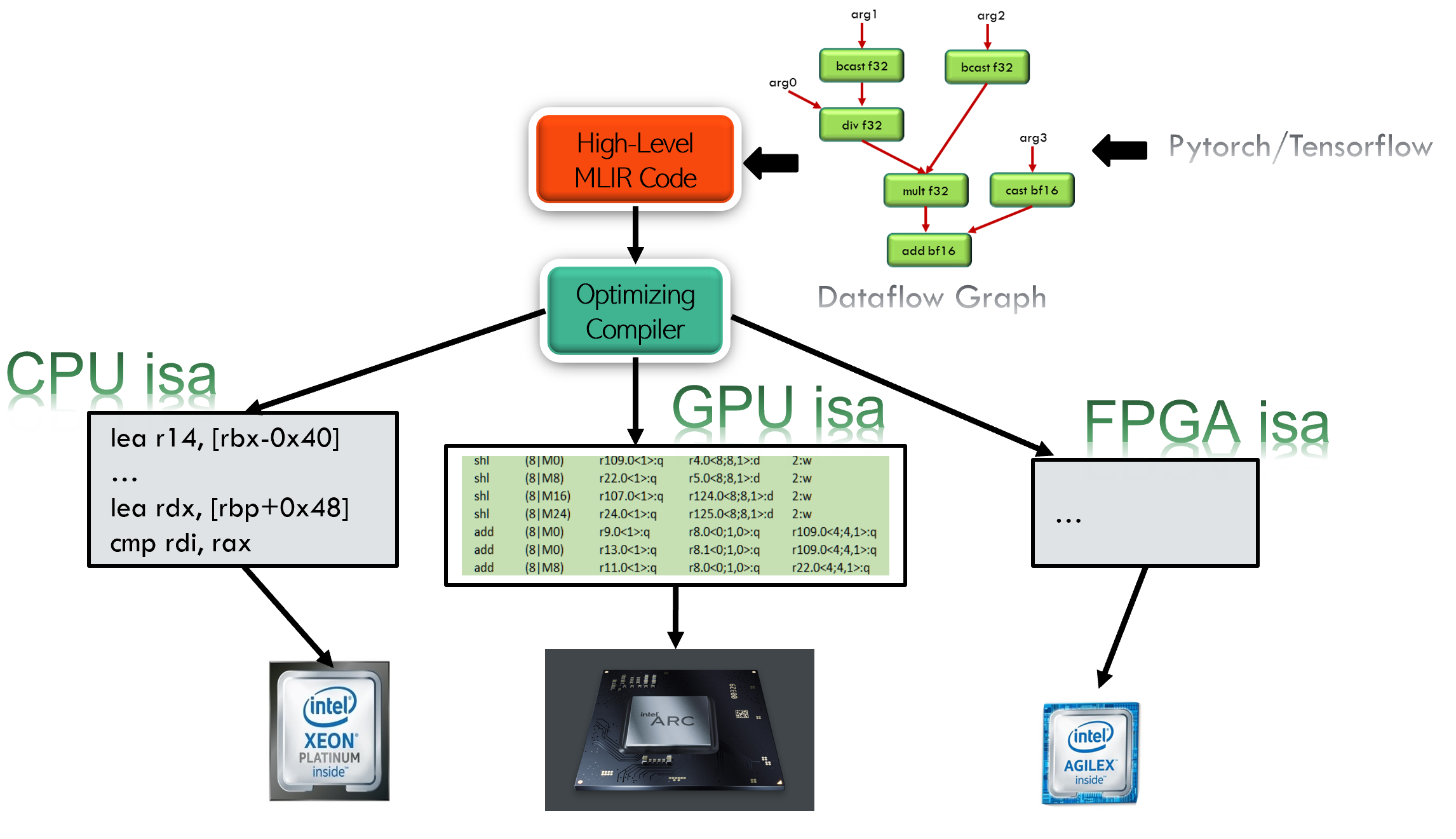}
  \end{center}
  \caption{DL-compiler flow}
  \label{fig:DLcompilerflow}
\end{figure}

In this paper, we apply ML/DL-models for estimating various hardware/machine characteristics for the benefit of compiler optimizations. Today's DL frameworks like Pytorch or Tensorflow generate dataflow graphs that encapsulate the various ML/DL models that data scientists and practitioners build. A Deep Learning (DL)-compiler then tries to $lower$ such a dataflow graph consisting of high-level operators and their data dependencies into actual machine code which runs on an xPU (CPU/GPU/TPU/AI accelerator etc). The flow of the DL-compiler from its dataflow graph input to the machine code output is complex and has many compiler optimization and lowering steps as shown in Fig ~\ref{fig:DLcompilerflow}. In order to make this task manageable and more democratized so that the compiler practitioners do not need to deal with a mushrooming of various IR (Intermediate Representation ) formats, MLIR ~\cite{mlir2021} was proposed in 2019. 

MLIR is part of the popular LLVM ~\cite{Lattner2004} toolchain and has been created to ease the translation of dataflow graphs emitted by Pytorch/Tensorflow and similar frameworks. Work on MLIR began with a realization that modern machine learning frameworks are composed of
many different compilers, graph technologies, and runtime systems — which do not
share a common infrastructure or design point, and not all of which follow best practices in compiler design. This manifested in multiple user-visible ways, including poor error messages, failures in edge cases, unpredictable performance, and difficulty generalizing the stack to support new hardware. Central to MLIR is the concept of $dialects$ which is a customizable IR instead of a one-size-fits-all IR like LLVM-IR. Depending on the framework being supported and/or the underlying hardware and various other restrictions, the compiler designer can build various MLIR $dialects$ with $progressive-lowering$ from one dialect to another. These dialects can co-exist with each other at the same time. The higher-level MLIR dialects are similar/close to the dataflow graph representations of the ML models. The lower-levels dialects are closer to LLVM-IR and machine ISA. While we lower higher level dialects to lower-level ones we apply various popular compiler as well as ML-specific optimizations ex: operator fusion, improved memory allocation for SW-controlled scratchpad memory, operator scheduling, loop interchange etc.

In order to efficiently apply compiler optimizations in MLIR form, the DL-compiler may need to have a fair estimate of the various hardware characteristics much before the actual code is generated ex: if we need to unroll a loop should we unroll-by-4 or an unroll-by-8 ? Do we run out of hardware resources like scratchpad memory or on-chip registers leading to register spilling when we unroll aggressively or fuse operators aggressively ? These and similar questions need to be answered by the DL-compiler as we generate code and lower progressively from a higher to a lower dialect. But to answer these questions effectively while the compilation is in progress inhibits compiling various versions and compare the parameters, else a very high compile time cost is incurred which a DL-compiler would like to avoid. Hence we would like to predict these hardware characteristics/parameters from a high level MLIR specification of the dataflow graph or even a lower level MLIR dialect like $affine$ without actually compiling and executing such a code end-to-end. In order to carry out this prediction effectively we visualize the MLIR code sequence as a sequence of text tokens as in popular NLP models \cite{Sutskever2014}. And then apply we following steps:
\begin{itemize}
    \item Tokenize the MLIR code using a vocabulary that encompasses the MLIR opcodes, various tensor keywords and tensor dimensions
    \item Use an embedding layer to convert the tokens to a dense vector representation
    \item Use a sequence-to-HW-characteristic regression model that inputs the dense vectors of the MLIR code as a sequence and outputs the said hardware characteristic ex: number of registers used or the latency of the MLIR code sequence
    \item Train such a model in a supervised manner
    \item Deploy the model which the DL-compiler can invoke while compiling in order to make the best decisions for effective lowering/code generation
\end{itemize}
\begin{figure*}[ht]
  \centering
   \includegraphics[width=0.95\linewidth]{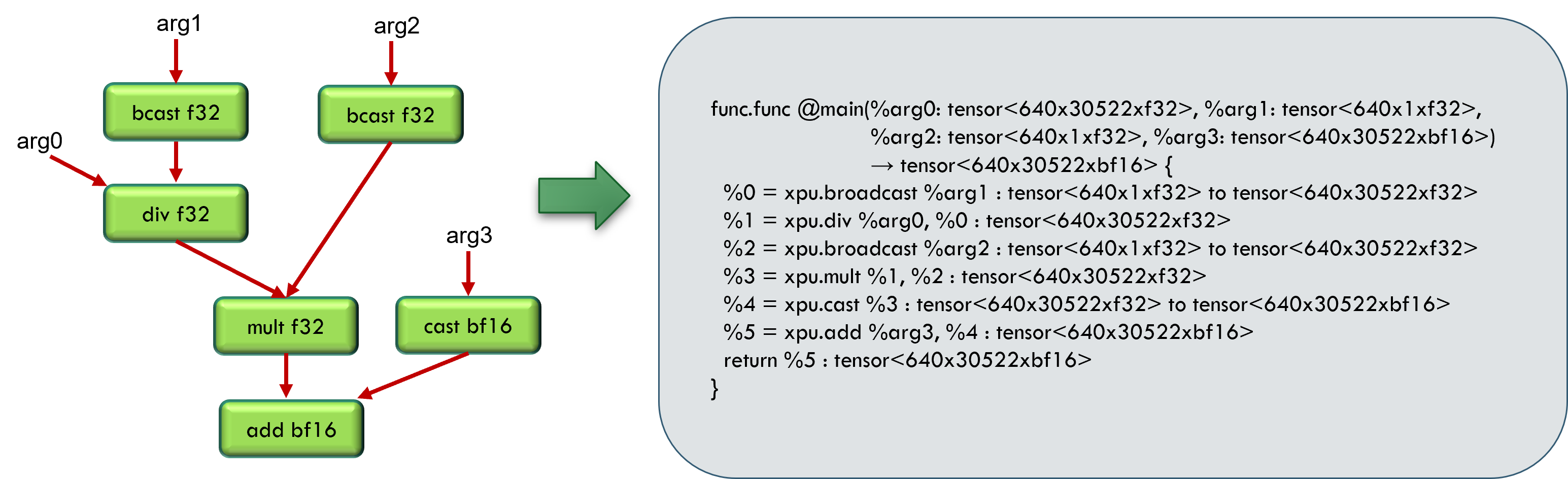}
  \caption{Dataflow Graph and its corresponding High-level MLIR}
  \label{fig:dataflow}
\end{figure*}
In the following sections we describe the structure of high-level MLIR that represents an ML dataflow graph. We provide details of the ML-based HW cost models we have
built for such an MLIR representation in order to predict two hardware characteristics - 1) Register Usage and 2) Latency/Cycles consumed for running the ML dataflow graph or subgraph.
We show some preliminary results on the accuracy of the ML model we have designed while predicting these hardware characteristics. We end with some comments on future work
and challenges.

\section{Background}
In order to understand the problem at hand, let us look at an example dataflow graph(subgraph) shown in Fig ~\ref{fig:dataflow} that has been generated via a framework like Pytorch/Tensorflow. In addition to the dataflow graph we show an equivalent MLIR representation of the graph that a DL-compiler will take as an input before applying various optimization and transformations. In the datflow graph each node is a high-level mathematical operator that needs to be lowered to the machine code of the corresponding hardware and the edges reflect the data dependencies that must be honored as the DL-compiler generates code for the entire graph(subgraph). The MLIR representation of this graph is a high-level textual IR representation, where the function embodies the graph and the operators appear as MLIR opcodes. The notation {\bf xpu} represents the name of the MLIR dialect which is xpu in this case and is a special dialect designed for our hardware. As an example, {\bf xpu.mult} represents the high-level multiplication operation that needs to be lowered by the DL-compiler. The data dependencies are modeled via the use-def chains where the defs are in SSA form \cite{SSA}. The DL ops in xpu dialect operate on a datatype called $tensor$ which are multi-dimensional matrices with a specified basic datatype.

\section{Machine-learnt Hardware Cost Model}

\begin{figure}[ht]
  \centering
   \includegraphics[width=0.8\linewidth]{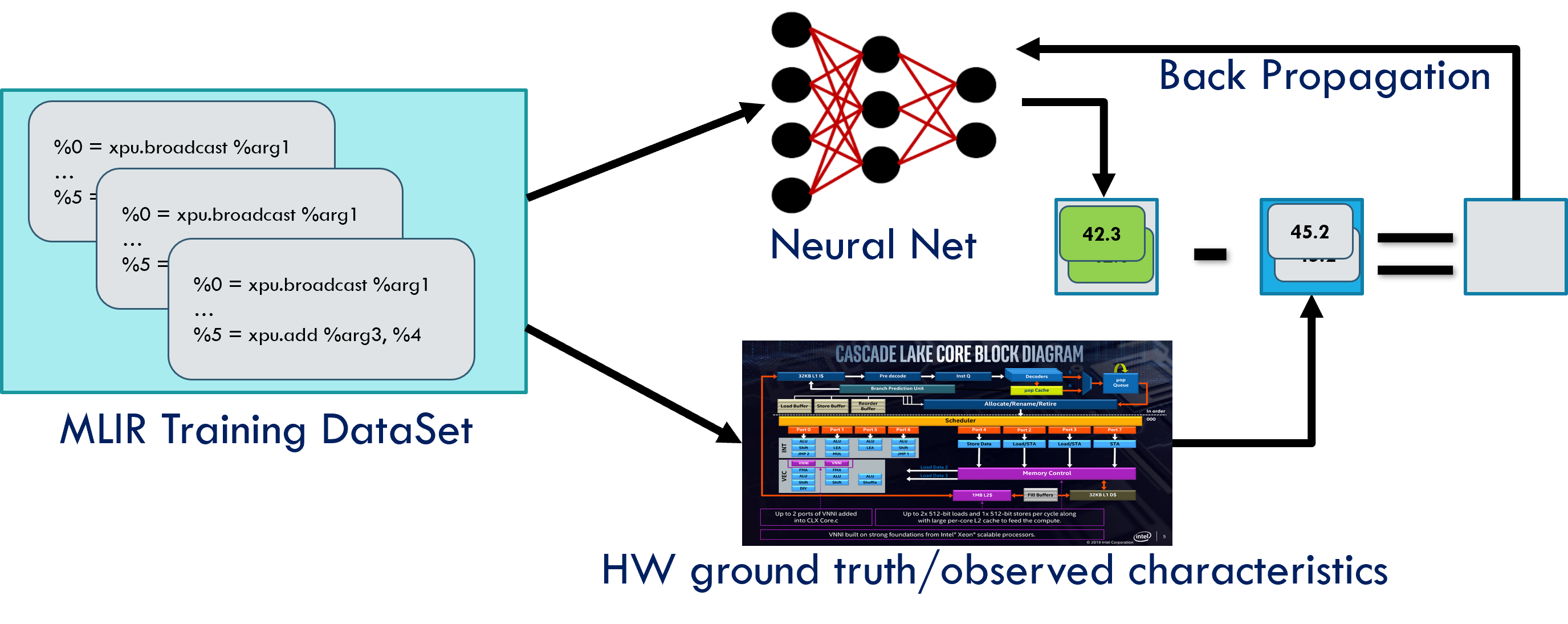}
  \caption{Overall ML-driven architecture}
  \label{fig:costmodelarch}
\end{figure}

Our overall ML-driven cost model prediction architecture is shown in Fig ~\ref{fig:costmodelarch}. The architecture is inspired by the work of Mendis et al. \cite{ithemal2019}. For training our cost model, we use a large dataset comprising of MLIR representations of dataflow graphs extracted form popular neural-net architectures like Resnet, BERT, Unet, SSD and Yolo. For this dataset we compile and run these graphs via the DL-compiler to capture the ground truth for the hardware characteristic to be predicted ex: register usage/pressure or latency/cycle count. Our ML-model ( described below ) is trained on the said dataset using the ground truth.

We will now describe in details the various aspects of our ML-model: 

{\bf Training Dataset:} Given a set of graphs (MLIR level functions), we want to create a dataset which can be fed to an ML model for a target variable prediction. We create a csv file for training consisting of : 1) Full MLIR Text sequence 2) Input and output tensor shapes 3) XPU utilization or register pressure as a target variable. Currently we have more than 20K MLIR files in the training set. In addition, we use augmentation to create a larger training set for better model training. 

\begin{figure*}[ht]
  \centering
   \includegraphics[width=1.0\linewidth]{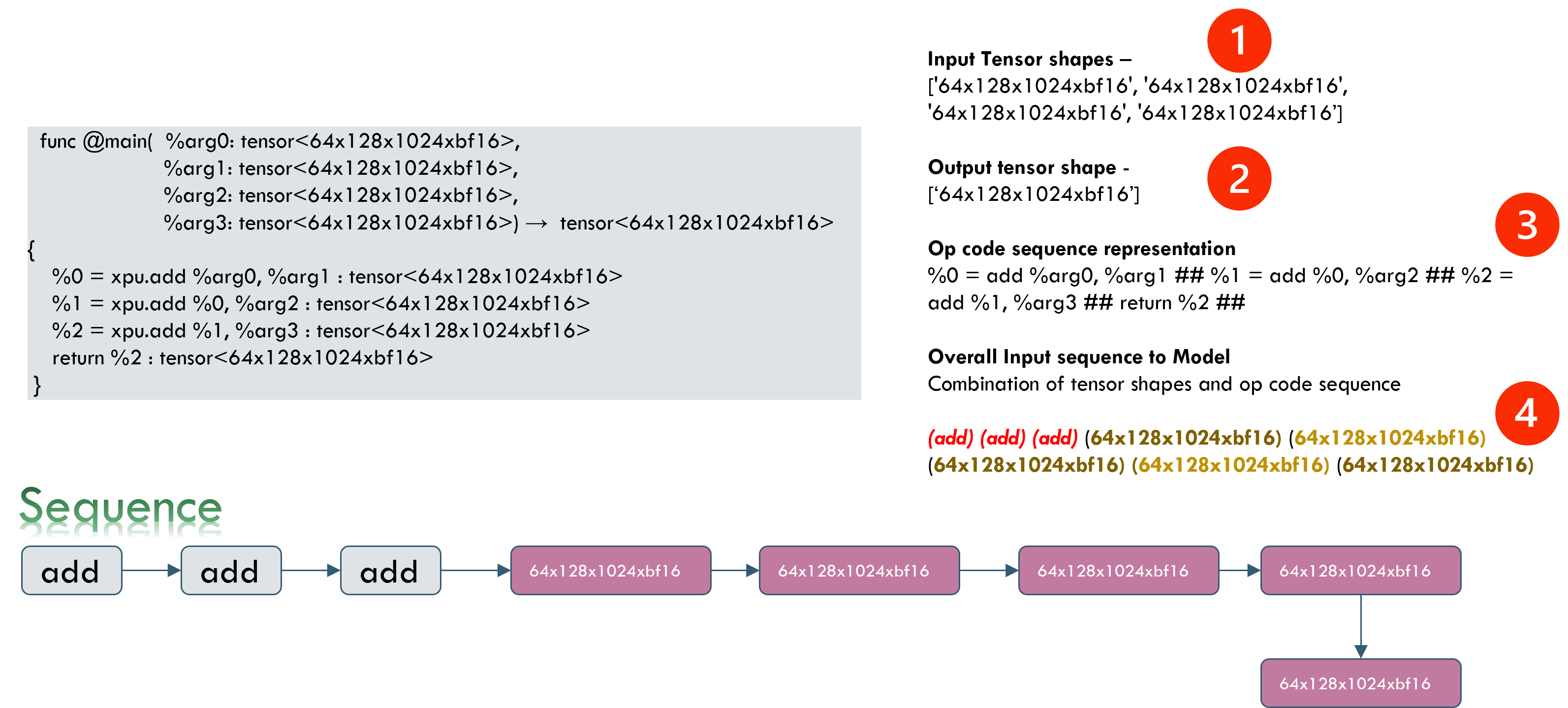}
  \caption{Tokenization and the Final Input sequence}
  \label{fig:tokenization}
\end{figure*}

{\bf Tokenization and Embedding: } We carry out two kinds of tokenization. In the first kind we just pick the xpu.op sequence and drop any other operand information which means we do not track the data dependence in this technique. Here, we tokenize the input and output tensor shapes as a single entity instead of breaking them down to their individual dimension values. This policy can result in some OOV tokens later but since in DL subgraphs many of the tensor sizes appear frequently across multiple models, the probability of OOV tokens remains low. We ensure that our training set encompasses most of the frequently used tensor shapes to have as few OOV tokens as possible. In the second kind of tokenization we maintain the xpu.ops as well as the operands as a sequence along with the tensor shapes. Such a sequence is usually up to 4x longer the op-only sequence. The sub-parts for the first kind of tokenization is demonstrated in Fig ~\ref{fig:tokenization}. The various pieces of tokenization is shown as sub-parts (1) to (4). The final sequence of tokens that is fed to the ML-model is shown at the bottom of the figure. Once the tokenization is done we pass the tokens via an embedding layer that creates a dense vector of dimension size 64 for each of the tokens. 

{\bf The Actual ML-model: } For our experiments we tried several ML models. These are: 1) A simple sequence of fully connected (FC) layers which considers the input token sequence as a bag-of-tokens 2) LSTM which ingests the input token sequence as-is and 3) Stacked Conv1D layers followed by MaxPool and FC which ingests the input token sequence as-is.

We observe that the first FC-based model has a high RMSE ( root mean square error ) while the LSTM-based model performs better than the first one on the RMSE front. However, it is the stacked Conv1D$+$MaxPool$+$FC model which performs the best with the lowest RMSE among the three models tried.
\begin{figure}[ht]
  \centering
   \includegraphics[width=0.5\linewidth]{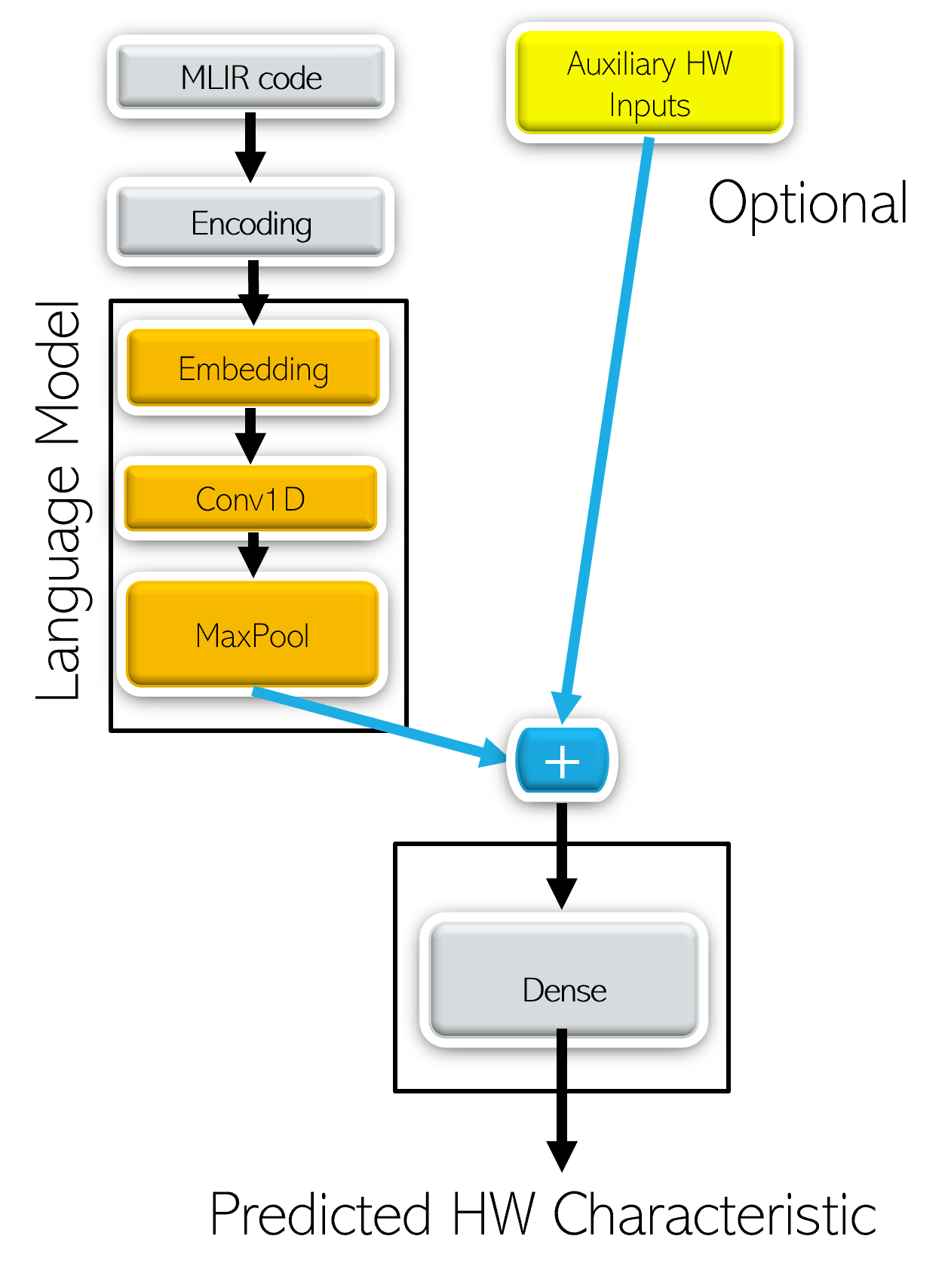}
  \caption{Conv1D+MaxPool+FC model}
  \label{fig:model}
\end{figure}
The particular model consists of 6 stacked Conv1D layers of filter size=2, a single MaxPool1D layer and 3 FC layers and predicts register Pressure/XPU Utilization. The overall structure of the model is given in Fig ~\ref{fig:model}.

\section{Experimental Evaluation}
In order to evaluate the performance and accuracy of our ML-based hardware cost model we use two hardware characteristics for our studies. The first is called $register pressure$ whereby we predict (from the high-level MLIR representation of dataflow graphs) the number of registers that the snippet of code will consume. This is important to evaluate as excessive register pressure can lead to spilling causing significant drop in performance during execution. The second hardware parameter that we predict is called $xpu utilization$. In this case, we evaluate how the hardware resources of the underlying xpu is utilized. For this particular study we concentrate on the hardware utilization of only the vector ALU unit which is prevalent in any modern xpu especially AI hardware accelerators. The utilization measures the number of times the vector ALU unit is utilized in a sequence of instructions. Higher the usage, higher is the utilization pointing to superior code generation by the DL compiler. 
\begin{figure*}[ht]
  \centering
   \includegraphics[width=1.0\linewidth]{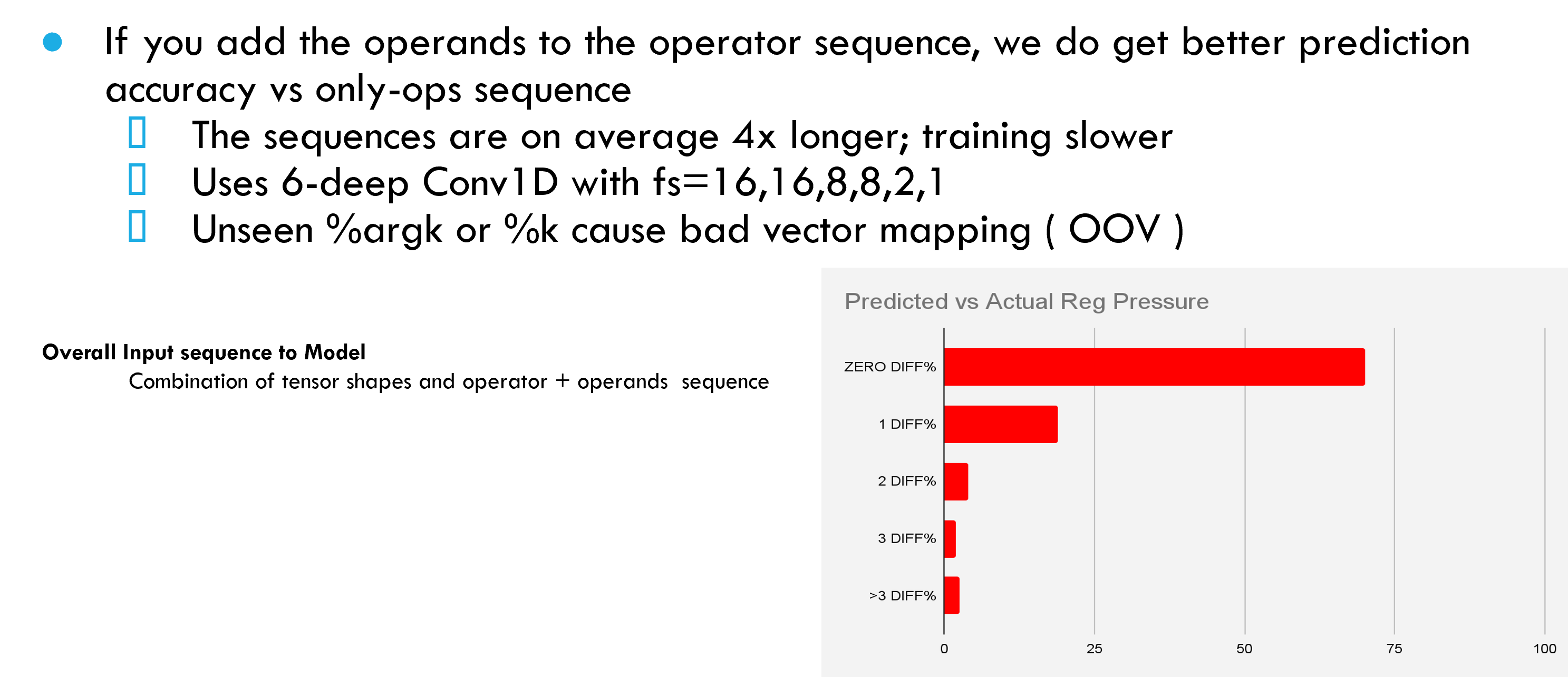}
  \caption{Results for Operator + Operand Modelling}
  \label{fig:opndmodel}
\end{figure*}
We use Intel's in-house DL-compiler and one of its major AI accelerators as the test bed. We train our model using 20k+ training samples as mentioned in Section 3. For these samples we gather the actual register pressure and xpu utilization factors by running these samples on the AI accelerator and collecting the relevant data which make up the ground truth. We use about 2K+ samples for inference/test. The RMSE values of the 2k+ test set, for both register pressure and xpu utilization studies, show that we get good accuracy and a small RMSE in the range of 5-7\%.

The results shown till now are for the case where only the xpu operators are tokenized ignoring the operands. We have also experimented with a longer sequence of tokens comprising both the xpu operators and its operands. In this case the accuracy improves further though we use an ML-model which is a slight modification of the one shown in Fig ~\ref{fig:model} where we use different filter sizes. The resultant structure and the accuracy are shown in Fig ~\ref{fig:opndmodel}. We can see that in almost 75\% of cases we can predict register pressure without any error.

\section{Related Work}
There has been some recent work on cost modelling of hardware and predicting certain hardware characteristics from high level code or even assembly. One of the earliest well-known work is a system called Ithemal by Mendis et al. ~\cite{ithemal2019} built to predict the latency/throughput of a block of assembly instructions for x86 CPUs. Ithemal used a two-layer hierarchical LSTM as a model to understand the x86 opcodes and their operands and finally output the latency/cycles of the basic block of assembly instructions. Following this work, Kaufman et al. built a ML-learnt performance model for Google Tensor Processing units (TPU) \cite{kaufman2021}. In this work the authors take a dataflow graph and try to predict the runtime of such a graph on TPU. The objective of the learnt model is to do better code generation on TPU. For this, they convert the dataflow graph to a Graph Neural Network (GNN) representation and train this GNN for final inferencing. Hunter et al. \cite{Hunter2022} builds a specific leant model for Intel VPUs called VPUNN to infer hardware performances.  They present ‘VPUNN’— a neural network-based cost model trained on low-level task profiling that consistently outperforms the state-of-the-art cost modeling in Intel’s line of VPU processors. Zhai et al. ~\cite{tlp2022} builds a DL-based cost model for Tensor Program tuning, but their approach is based on feature selection of the tensor programs similar to AutoTVM ~\cite{autotvm2019}. Similarly, Baghdadi et al. develop a DL-based cost model for tensor programs written in a DSL called Tiramisu ~\cite{baghdadi2021}. However the cost-model is based on hand tuned features and is unlikely to be scalable like the other previous efforts.
Though slightly orthogonal to the current work, Cummins et al.'s work to learn a compiler heuristic end-to-end for mapping OpenCL kernels effectively to a CPU/GPU has also several interesting proposals and techniques that can be effectively utilized for predicting hardware performance characteristics.

Our work is the first of its kind which tackles hardware performance or bottleneck predictions for MLIR code. The model developed by us based on Conv1D and Maxpool is an extremely fast and accurate model compared to the likes of LSTM or even GNNs. In addition, our model is scalable to different forms of MLIR - from high-level MLIR dialects to lower-level dialects like $affine$ or $scf$ which can produce much larger sequences of the order of thousands of tokens due to the presence of loops and control flow. Finally, model works with the state-of-the-art DL compiler infrastructure involving MLIR and LLVM.
 
\section {Conclusion and Future Work}
In this paper, we have outlined a mechanism to infer/project performance characteristics of the underlying hardware from MLIR code that helps a DL-compiler take actions that lead to better code generation. Our technique is based on viewing the MLIR code as a text sequence and applying well-known NLP-like models for this purpose. We show good accuracy in inferring such hardware performance characteristics though the code is quite removed from the actual hardware ISA that the DL compiler finally generates. We envision such methods to be of use even for standard non-DL compilers and can replace popular hardware predictors like LLVM's TTI interface.

As part of future work we envision several things - 1) Use more powerful models like Transformers to better the currently achieved accuracy figures 2) Use larger training sets to reduce OOV errors 3) Integrate the inference infrastructure ( which is currently standalone ) in a regular DL-compiler flow and 4) Use such a technique for more complex hardware characteristic prediction like cache misses and memory traffic projection.

One of the challenges we see is with respect to prediction of actual runtimes as the universe of tensor sizes on which the runtime depends encompasses the natural number set. Tokenizing natural numbers accurately is a problem in ML due to which we see predictions of runtime estimates having wider prediction variability compared to other hardware characteristics. This is an area that needs more active research.

\bibliography{hwcostmodel}
\bibliographystyle{iclr2023_conference}
\end{document}